\documentclass[10pt,twocolumn,letterpaper]{article}

\usepackage{iccv}
\usepackage{times}
\usepackage{epsfig}
\usepackage{graphicx}
\usepackage{amsmath}
\usepackage{amssymb}
\usepackage{times}
\usepackage{epsfig}
\usepackage{graphicx}
\usepackage{amsmath}
\usepackage{amssymb}
\usepackage{multirow}
\usepackage{booktabs} % Allows the use of \toprule, \midrule and
\usepackage{algorithm}
\usepackage{algorithmic}
\usepackage{subfigure}
\usepackage{balance}
\usepackage[pagebackref=true,breaklinks=true,letterpaper=true,colorlinks,bookmarks=false]{hyperref}

\usepackage{color}

\iccvfinalcopy % *** Uncomment this line for the final submission

\def\iccvPaperID{198} % *** Enter the ICCV Paper ID here

\ificcvfinal\fi
\begin{document}
	
	%%%%%%%%% TITLE
	\title{Recurrent Topic-Transition GAN for Visual Paragraph Generation}
	
	\author{Xiaodan Liang\\
Carnegie Mellon University\\
{\tt\small xiaodan1@cs.cmu.edu}
\and
Zhiting Hu\\
Carnegie Mellon University\\
{\tt\small zhitingh@cs.cmu.edu}
\and
Hao Zhang\\
Carnegie Mellon University\\
{\tt\small hao@cs.cmu.edu}
\and
Chuang Gan\\
Tsinghua University\\
{\tt\small ganchuang1990@gmail.com}
\and
Eric P. Xing\\
Carnegie Mellon University\\
{\tt\small epxing@cs.cmu.edu}\\
}
	
	\maketitle
	%\thispagestyle{empty}

	%%%%%%%%% ABSTRACT
	\begin{abstract}
		A natural image usually conveys rich semantic content and can be viewed from different angles. Existing image description methods are largely restricted by small sets of biased visual paragraph annotations, and fail to cover rich underlying semantics. In this paper, we investigate a semi-supervised paragraph generative framework that is able to synthesize diverse and semantically coherent paragraph descriptions by reasoning over local semantic regions and exploiting linguistic knowledge. The proposed Recurrent Topic-Transition Generative Adversarial Network (RTT-GAN) builds an adversarial framework between a structured paragraph generator and multi-level paragraph discriminators. The paragraph generator generates sentences recurrently by incorporating region-based visual and language attention mechanisms at each step. The quality of generated paragraph sentences is assessed by multi-level adversarial discriminators from two aspects, namely, plausibility at sentence level and topic-transition coherence at paragraph level. The joint adversarial training of RTT-GAN drives the model to generate realistic paragraphs with smooth logical transition between sentence topics. Extensive quantitative experiments on image and video paragraph datasets demonstrate the effectiveness of our RTT-GAN in both supervised and semi-supervised settings. Qualitative results on telling diverse stories for an image also verify the interpretability of RTT-GAN.		
	\end{abstract}
	
	\section{Introduction}
	Describing visual content with natural language is an emerging interdisciplinary problem at the intersection of computer vision, natural language processing, and artificial intelligence. Recently, great advances~\cite{kulkarni2011baby,chen2015mind,donahue2015long,yang2016review,mao2014deep,you2016image} have been achieved in describing images and videos using a single high-level sentence, owing to the advent of large datasets~\cite{lin2014microsoft,young2014image,krishna2016visual} pairing images with natural language descriptions. 
	Despite the encouraging progress in image captioning~\cite{xu2015show,you2016image,mao2014deep,yang2016review}, most current systems tend to capture the scene-level gist rather than fine-grained entities, which largely undermines their applications in real-world scenarios such as blind navigation, video retrieval, and automatic video subtitling. One of the recent alternatives to sentence-level captioning is visual paragraph generation~\cite{Visual_story,krause2016hierarchical,yu2016video,xu2016msr}, which aims to provide a coherent and detailed description, like telling stories for images/videos.
	
	\begin{figure}[!tp]
		\begin{center}
			\includegraphics[scale=0.52]{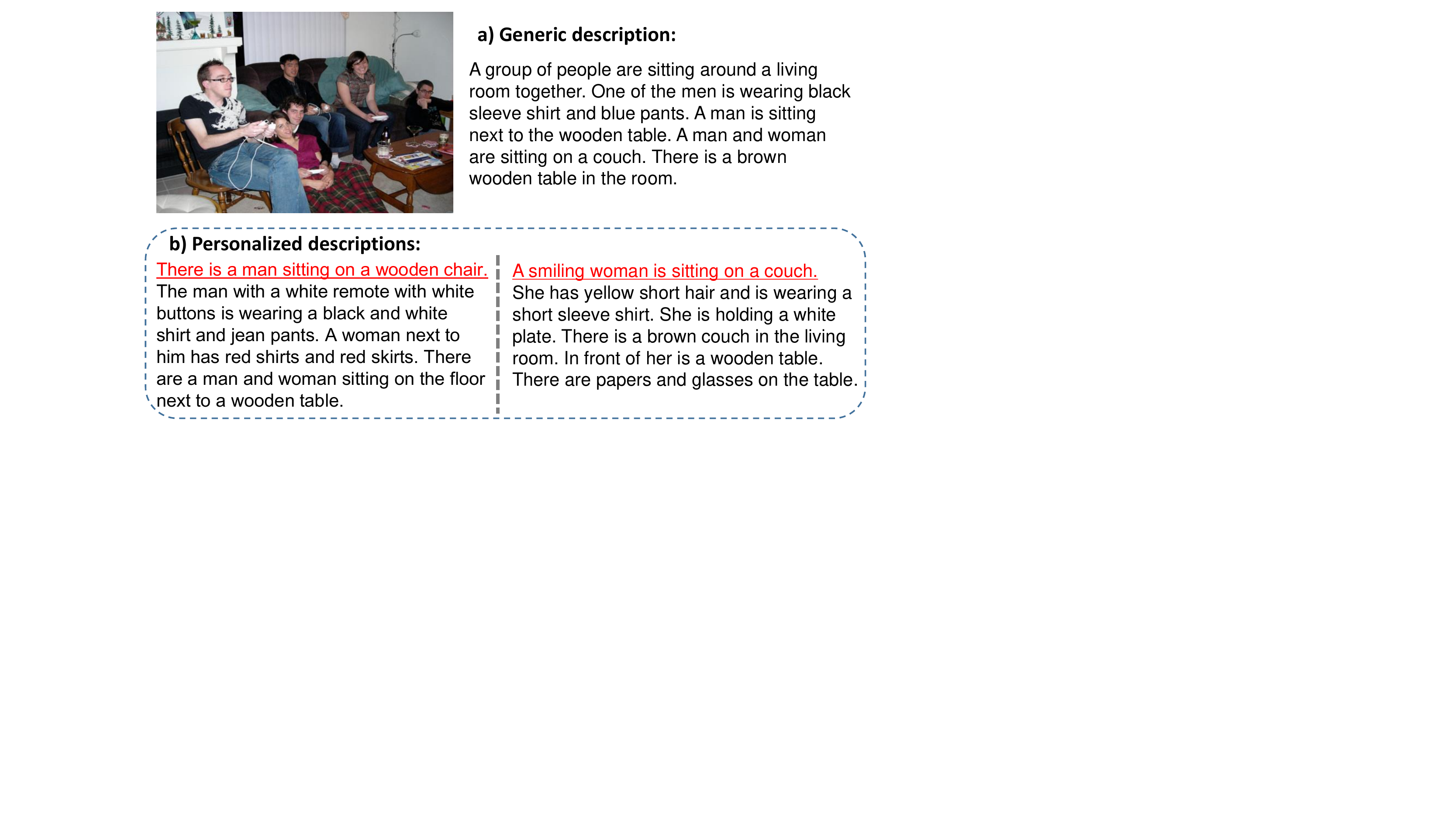}
			\caption{Our RTT-GAN is able to automatically produce generic paragraph descriptions shown in (a), and personalized descriptions by manipulating first sentences (highlighted in red), shown in (b).} 
			\label{fig:intuition}
		\end{center}
		\vspace{-4mm}
	\end{figure}
	
	Generating a full paragraph description for an image/video is challenging. First, paragraph descriptions tend to be diverse, just like different individuals can tell stories from personalized perspectives. As illustrated in Figure~\ref{fig:intuition}, users may describe the image starting from different viewpoints and objects. Existing methods~\cite{krause2016hierarchical,yu2016video,li2015hierarchical} deterministically optimizing over single annotated paragraph thus suffer from losing massive information expressed in the image. It is desirable to enable diverse generation through simple manipulations.
	Second, annotating images/videos with long paragraphs is labor-expensive, leading to only small scale image-paragraph pairs which limits the model generalization. 
	Finally, different from single-sentence captioning, visual paragraphing requires to capture more detailed and richer semantic content. It is necessary to perform long-term visual and language reasoning to incorporate fine-grained cues while ensuring coherent paragraphs.

	To overcome the above challenges, we propose a semi-supervised visual paragraph generative model, Recurrent Topic-Transition GAN (RTT-GAN), which generates diverse and semantically coherent paragraphs by reasoning over both local semantic regions and global paragraph context. 
	Inspired by Generative Adversarial Networks (GANs)~\cite{goodfellow2014generative}, we establish an adversarial training mechanism between a structured paragraph generator and multi-level paragraph discriminators, where the discriminators learn to distinguish between real and synthesized paragraphs while the generator aims to fool the discriminators by generating diverse and realistic paragraphs.
	
	The paragraph generator is built upon dense semantic regions of the image, and selectively attends over the regional content details to construct meaningful and coherent paragraphs. To enable long-term visual and language reasoning spanning multiple sentences, the generator recurrently maintains context states of different granularities, ranging from paragraph to sentences and words. Conditioned on current state, a spatial visual attention mechanism selectively incorporates visual cues of local semantic regions to manifest a topic vector for next sentence, and a language attention mechanism incorporates linguistic information of regional phrases to generate precise text descriptions. 
	We pair the generator with rival discriminators which assess synthesized paragraphs in terms of plausibility at sentence level as well as topic-transition coherence at paragraph level. 
	Our model allows diverse descriptions from a single image by manipulating the first sentence which guides the topic of the whole paragraph. Semi-supervised learning is enabled in the sense that only {\it single-sentence} caption annotation is required for model training, while the linguistic knowledge for constructing long paragraphs is transfered from standalone text paragraphs without paired images.

We compare RTT-GAN with state-of-the-art methods on both image-paragraph and video-paragraph datasets, and verify the superiority of our method in both supervised and semi-supervised settings. Using only the single-sentence COCO captioning dataset, our model generates highly plausible multi-sentence paragraphs. 
%In terms of semi-supervised learning, our RTT-GAN enables to generate reasonable whole paragraph when only the singe-sentence captioning in large-scale COCO captioning dataset is provided. 
Given these synthesized paragraphs for COCO image, we can considerably enlarge the existing small paragraph dataset to further improve the paragraph generation capability of our RTT-GAN. 
Qualitative results on personalized paragraph generation also shows the flexibility and applicability of our model.
	
	\section{Related Work}
	
	\textbf{Visual Captioning.} Image captioning is posed as a longstanding and holy-grail goal in computer vision, targeting at bridging visual and linguistic domain. Early works that posed this problem as a ranking and template retrieval tasks~\cite{farhadi2010every,hodosh2013framing,karpathy2014deep} performed poorly since it is hard to enumerate all possibilities in one collected dataset due to the compositional nature of language. Therefore, some recent works~\cite{kulkarni2011baby,chen2015mind,donahue2015long,yang2016review,mao2014deep,you2016image,liang2017deep} focus on directly generating captions by modeling the semantic mapping from visual cues to language descriptions. Among all these research lines, advanced methods that train recurrent neural network language models conditioned on image features~\cite{chen2015mind,donahue2015long,yang2016review,you2016image} achieve great success by taking advantages of large-scale image captioning dataset. Similar success has been already seen in video captioning fields~\cite{donahue2015long,yao2015describing}. Though generating high-level sentences for images is encouraging, massive underlying information, such as relationships between objects, attributes, and entangled geometric structures conveyed in the image, would be missed if only summarizing them with a coarse sentence. Dense captioning~\cite{johnson2016densecap} is recently proposed to describe each region of interest with a short phrase, considering more details than standard image captioning. However, local phrases can not provide a comprehensive and logical description for the entire image.

	\textbf{Visual Paragraph Generation.} Paragraph generation overcomes shortcomings of standard captioning and dense captioning by producing a coherent and fine-grained natural language description.  To reason about long-term linguistic structures with multiple sentences, hierarchical recurrent network~\cite{li2015hierarchical,lin2015hierarchical,yu2016video,krause2016hierarchical} has been widely used to directly simulate the hierarchy of language. For example, %Li et al.~\cite{li2015hierarchical} adopt a hierarchical autoencoder. while Lin et al.~\cite{lin2015hierarchical} use different recurrent units to model sentences and words. 
	Yu et al.~\cite{yu2016video} generate multi-sentence video descriptions for cooking videos to capture strong temporal dependencies. Krause et al.~\cite{krause2016hierarchical} combine semantics of all regions of interest to produce a generic paragraph for an image. However, all these methods suffer from the overfitting problem due to the lack of sufficient paragraph descriptions. In contrast, we propose a generative model to automatically synthesize a large amount of diverse and reasonable paragraph descriptions by learning the implicit linguistic interplay between sentences. Our RTT-GAN has better interpretability by imposing the sentence plausibility and topic-transition coherence on the generator with two adversarial discriminators. The generator selectively incorporates visual and language cues of semantic regions to produce each sentence.
	
		\begin{figure*}[!tp]
			\begin{center}
				\includegraphics[scale=0.7]{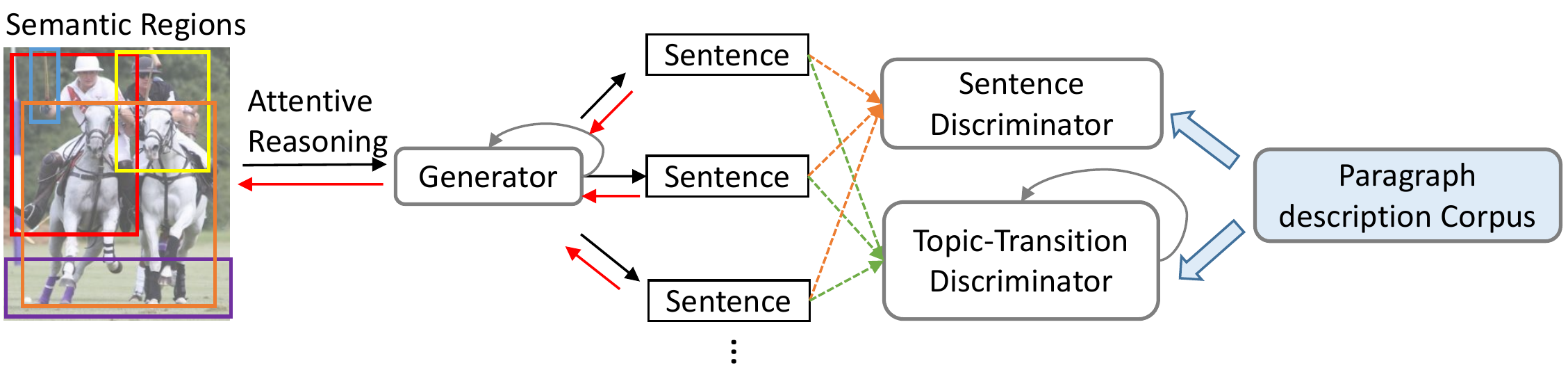}
				\caption{Our RTT-GAN alternatively optimizes a structured paragraph generator and two discriminators following an adversarial training scheme. The generator recurrently produces each sentence by reasoning about local semantic regions and preceding paragraph state. Each synthesized sentence is then fed into a sentence discriminator and a recurrent topic-transition discriminator for assessing sentence plausibility and topic coherence, respectively. A paragraph description corpus is adopted to provide linguistic knowledge about paragraph generation, which depicts the true data distribution of the discriminators .} 
				\label{fig:framework}
			\end{center}
			\vspace{-2mm}
		\end{figure*}
		
	\section{Recurrent Topic-Transition GAN}
	The proposed Recurrent Topic-Transition GAN (RTT-GAN) aims to describe the rich content of a given image/video by generating a natural language paragraph. Figure~\ref{fig:framework} provides an overview of the framework. Given an input image, we first detect a set of semantic regions using dense captioning method~\cite{johnson2016densecap}. Each semantic region is represented with a visual feature vector and a short text phrase (e.g. {\it person riding a horse}).
	The paragraph generator then sequentially generates meaningful sentences by incorporating the fine-grained visual and textual cues in a selective way. 
	%through tailored attention mechanisms.
	To ensure high-quality individual sentences and coherent whole paragraph, we apply a sentence discriminator and a topic-transition discriminator on each generated sentence, respectively, to measure the plausibility and smoothness of semantic transition with preceding sentences. The generator and multi-level discriminators are learned jointly within an adversarial framework. RTT-GAN supports not only supervised setting with annotated image-paragraph pairs, but also semi-supervised setting where only a single sentence caption is provided for each image and the knowledge of long paragraph construction is learned from a standalone paragraph corpus.
	
	In next sections, we first derive the adversarial framework of our RTT-GAN, then describe detailed model design of the paragraph generator and the multi-level discriminators, respectively. 
	
	\begin{figure*}[!tp]
		\begin{center}
			\includegraphics[scale=0.48]{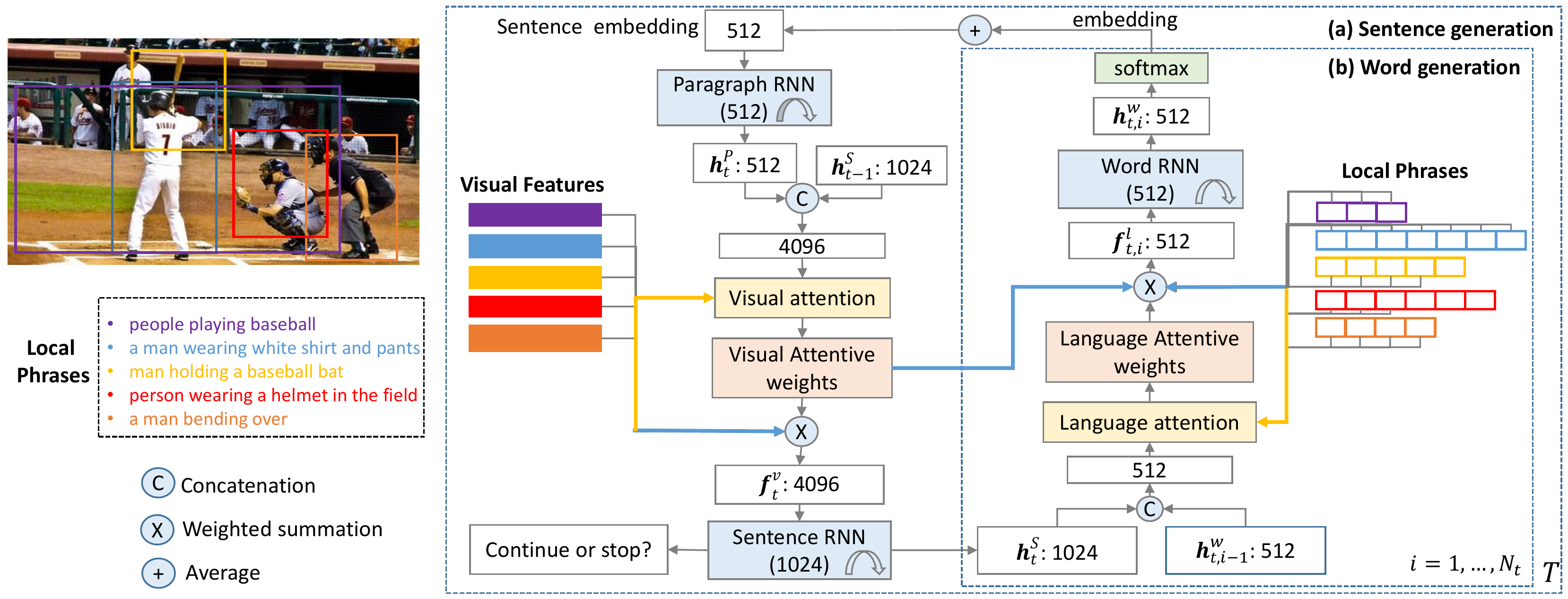}
			\caption{Illustration of our paragraph generator. Given visual features and local phrases of semantic regions, the paragraph generator is performed for most $T$ steps to sequentially generate each sentence. At $t$-th step, the paragraph states $\mathbf{h}_t^P$ is first updated with the embedding of preceding sentences by \emph{paragraph RNN}. Then, the visual attention takes features of semantic regions, current paragraph states $\mathbf{h}_t^P$ and previous hidden states $\mathbf{h}_{t-1}^S$ as input to manifest a visual context vector $\mathbf{f}_t^v$. $\mathbf{f}_t^v$ is then fed into \emph{sentence RNN} to obtain the encoded topic vector $\mathbf{h}_{t}^S$ and determine whether to generate next sentence. The \emph{word RNN} with language attention then generates each word. } 
			\label{fig:generator}
		\end{center}
		\vspace{-6mm}
	\end{figure*}
	
\subsection{Adversarial Objective}\label{sec:objective}
We construct an adversarial game between the generator and discriminators to drive the model learning. Specifically, the sentence and topic-transition discriminators learn a critic between real and generated samples, while the generator attempts to confuse the discriminators by generating realistic paragraphs that satisfy linguistic characteristics (i.e., sentence plausibility and topic-transition coherence). The generative neural architecture ensures the paragraph captures adequate semantic content of the image, which we describe in detail in the next sections. Formally, let $G$ denote the paragraph generator, and let $D^s$ and $D^r$ denote the sentence and topic-transition discriminators, respectively. 
%In this section, we will formally derive the training objective of our RTT-GAN. The vanilla Generative adversarial nets (GANs)~\cite{goodfellow2014generative} proposes simultaneously trains the generator $G$ and discriminator $D$ so that $G$ learns to generate samples from random noises that are hard to classify by $D$, while $D$ learns to discriminate the samples generated by $G$. In the spirit of GANs, RTT-GAN learns to generate realistic paragraphs for an image that satisfy the natural linguistic characteristics (i.e., sentence plausibility and topic-transition coherence) by solving the adversarial min-max problem. Formally, RTT-GAN optimizes one recurrent paragraph generators $G$, and paragraph discriminators $D^p$ including a sentence discriminator $D^s$ and recurrent topic-transition discriminator $D^r$. 

%
At each time step $t$, conditioned on preceding sentences $\mathbf{s}_{1:t-1}$ and local semantic regions $\mathbf{V}$ of the image, the generator $G$ recurrently produces a single sentence $\mathbf{s}_t$, where each sentence $\mathbf{s}_t = \{\mathbf{w}_{t, i}\}$ consists of a sequence of $\mathbf{N}_t$ words $\mathbf{w}_{t, i}\  (i=1,\dots, \mathbf{N}_t)$:
\begin{equation}
\small
\begin{split}
P_{G}(\mathbf{s}_t| \mathbf{s}_{1:t-1}, \mathbf{V}) = \prod_{i=1}^{\mathbf{N}_t} P_{G}(\mathbf{w}_{t,i}| \mathbf{w}_{t,1:i-1}, \mathbf{s}_{1:t-1}, \mathbf{V}).
\label{global}
\end{split}
\end{equation}

The discriminators learn to differentiate real sentences $\hat{\mathbf{s}}$ within a true paragraph $\hat{\mathbf{P}}$ from the synthesized ones $\mathbf{s}_t$. The generator $G$ tries to generate realistic visual paragraph by minimizing against the discriminators' chance of correctly telling apart the sample source. 
%As discussed in the most recent Wasserstein GAN~\cite{arjovsky2017wasserstein}, 
As the original GAN~\cite{goodfellow2014generative} that optimizes over binary probability distance suffers from mode collapse and instable convergence, we follow the new Wasserstein GAN~\cite{arjovsky2017wasserstein} method that theoretically remedies this by minimizing an approximated Wasserstein distance. The objective of the adversarial framework is thus written as: 
% an adversarial optimization:
\begin{equation}
\begin{split}
&\min_{G}\max_{D^s, D^r}  \mathcal{L}(G,D^s, D^r) =\\
& \mathbb{E}_{\hat{\mathbf{s}} \sim p_{\text{data}(\hat{\mathbf{s}})}} \big[D^s(\hat{\mathbf{s}})\big]
- \mathbb{E}_{\mathbf{s}_{1:t} \sim p_{G(\mathbf{s}_{1:t} | \mathbf{V})}} \big[D^s(\mathbf{s}_t)\big] + \\ 
& \mathbb{E}_{\hat{\mathbf{P}} \sim p_{\text{data}(\hat{\mathbf{P}})}} \big[D^r (\hat{\mathbf{P}})\big]
- \mathbb{E}_{\mathbf{s}_{1:t} \sim p_{G(\mathbf{s}_{1:t}| \mathbf{V})}} \big[D^r(\mathbf{s}_{1:t})\big],
\label{eq:d-opt}
\end{split} 
\end{equation}
where $p_{\text{data}(\hat{\mathbf{s}})}$ and $p_{\text{data}(\hat{\mathbf{P}})}$ denote the true data distributions of sentences and paragraphs, respectively, which are empirically constructed from a paragraph description corpus. The second line of the equation is the objective of the sentence discriminator $D^s$ that optimizes a critic between real/fake sentences, while the third line is the objective of the topic-transition discriminator $D^r$. Here $p_{G(\mathbf{s}_{1:t}| \mathbf{V})}$ indicates the distribution of generated sentences by the generator $G$. 

To leverage existing image-paragraph pair dataset in the supervised setting, or image captioning dataset in the semi-supervised setting, we also incorporate the traditional word reconstruction loss for generator optimization, which is defined as:
\begin{equation}
\small
\begin{split}
\mathcal{L}^c(G) = -\sum_{t=1}^{T}\sum_{i=1}^{\mathbf{N}_t} \log P_{G}(\mathbf{w}_{t,i}|  \mathbf{w}_{t,1:i-1}, \mathbf{s}_{1:t-1}, \mathbf{V}).
\end{split}
\label{eq:lcg}
\end{equation}
%The generator $G$ is parametrized with a recurrent network that produces one sentence of a paragraph at each step. $G(\mathbf{s}_t|\mathbf{s}_{1:t-1},V)$ computes the generated sentence in each recurrent step, and $G(\mathbf{s}_{1:t} | V)$ denotes the generated paragraph composed by a sequence of sentences. 
Note that the reconstruction loss is only used for supervised examples with paragraph annotations, and semi-supervised examples with single-sentence caption (where we set $T=1$). Combining Eqs.\eqref{eq:d-opt}-\eqref{eq:lcg}, the joint objective for the generator $G$ is thus:
\begin{equation}
\begin{split}
G^* = \arg\min_G\max_{D^s, D^r} \lambda\mathcal{L}(G,D^s, D^r) + \mathcal{L}^{c}(G),
\end{split}
\label{eq:g-opt}
\end{equation}
where $\lambda$ is the balancing parameter fixed to $0.001$ in our implementation. The optimization of the generator and discriminators (i.e., Eq.\eqref{eq:g-opt} and Eq.\eqref{eq:d-opt}, respectively) is performed in an alternating min-max manner. We describe the training details in section~\ref{sec:imp}.

The discrete nature of text samples hinders gradient back-propagation from the discriminators to the generator~\cite{hu2017controllable}. We address this issue following SeqGAN~\cite{yu2016seqgan}. The state is the current produced words and sentences, and the action is the next word to select. we apply Monte Carlo search with a roll-out policy to sample the remaining words until it sees an END token for each sentence and maximal number of sentences. The roll-out policy is the same with the generator, elaborated in Section~\ref{sec:generator}.  The discriminator is trained by providing true paragraphs from the text corpus and synthetic ones from the generator.  The generator is updated by employing a policy gradient based on the expected reward received from the discriminator and the reconstruction loss for fully-supervised and semi-supervised settings, defined in Eq.~\ref{eq:g-opt}. To reduce the variance of the action values, we run the roll-out policy starting from current state till the end of the paragraph for five times to
get a batch of output samples. The signals that come from the word prediction for labeled sentences (defined in Eq.~\ref{eq:lcg})) can be regarded as the intermediate reward. The gradients are passed back to the intermediate action value via Monte Carlo search~\cite{yu2016seqgan}.

\subsection{Paragraph Generator}
\label{sec:generator}
Figure~\ref{fig:generator} shows the architecture of the generator $G$, which recurrently retains different levels of context states with a hierarchy constructed by a paragraph RNN, a sentence RNN, and a word RNN, and two attention modules. First, the paragraph RNN encodes the current paragraph state based on all preceding sentences. Second, the spatial visual attention module selectively focuses on semantic  regions with the guidance of current paragraph state to produce the visual representation of the sentence. The sentence RNN is thus able to encode a topic vector for the new sentence. Third, the language attention module learns to incorporate linguistic knowledge embedded in local phrases of focused semantic regions to facilitate word generation by the word RNN.  

	\textbf{Region Representation. } 
Given an input image, we adopt the dense captioning model~\cite{johnson2016densecap,krause2016hierarchical} to detect semantic regions of the image and generate their local phrases. Each region $\mathbf{R}_j\  (j\in 1, \dots, M)$ has a visual feature vector $\mathbf{v}_j$ and a local text phrase (i.e., region captioning) $\mathbf{s}^r_j = \{\mathbf{w}^r_{j,i}\}$ consisting of $\mathbf{N}_j$ words. In practice, we use the top $M=50$ regions.
	
	\textbf{Paragraph RNN. }
The paragraph RNN keeps track of the paragraph state by summarizing preceding sentences. 
%It is a single-layer LSTM with hidden size of 512 and initial hidden and memory cells set to zero. 
At each $t$-th step ($t = 1, \dots, T$), the paragraph RNN takes the embedding of generated sentence in previous step as input, and in turn produces the paragraph hidden state $\mathbf{h}^P_t$. The sentence embedding is obtained by simply averaging over the embedding vectors of the words in the sentence. This strategy enables our model to support the manipulation of the first sentence to initialize the paragraph RNN and generate personalized follow-up descriptions.
	
	\textbf{Sentence RNN with Spatial Visual Attention. }
The visual attentive sentence RNN controls the topic of the next sentence $\mathbf{s}_t$ by selectively focusing on relevant regions of the image. Specifically, given the paragraph states $\mathbf{h}^P_{t}$ from the paragraph RNN and previous hidden states $\mathbf{h}^S_{t-1}$ of the sentence RNN, we apply an attention mechanism on the visual features $\mathbf{V} = \{\mathbf{v}_1, \dots, \mathbf{v}_M\}$ of all semantic regions, and construct a visual context vector $\mathbf{f}^v_t$ that represents the next sentence at $t$-th step: 
\begin{equation}
\begin{split}
\mathbf{f}^v_t &= \text{att}^v(\mathbf{V}, \mathbf{h}^P_{t}, \mathbf{h}^S_{t-1})\\
&= \sum_{j=1}^{M}\frac{\alpha(\mathbf{v}_j, \beta(\mathbf{h}^P_{t}, \mathbf{h}^S_{t-1}))}{\sum_{j'=1}^{M}\alpha(\mathbf{v}_{j'}, \beta(\mathbf{h}^P_{t}, \mathbf{h}^S_{t-1}))}\mathbf{v}_j \\
&:= \sum_{j=1}^{M} a_j \mathbf{v}_j,
\end{split}
\end{equation}
where $\beta(\mathbf{h}^P_{t}, \mathbf{h}^S_{t-1})$ is a linear layer that transforms the concatenation of $\mathbf{h}^P_{t}$ and $\mathbf{h}^S_{t-1}$ into a compact vector with the same dimension as $\mathbf{v}_j$; the function $\alpha(\cdot)$ is to compute the weight of each region and is implemented with a single linear layer. 
%following~\cite{luong2015effective}. 
For notational simplicity, we use $a_j$ to denote the normalized attentive weight of each region $\mathbf{R}_j$.

Given the visual representation $\mathbf{f}^v_t$, the sentence RNN is responsible for determining the number of sentences that should be in the paragraph and producing a topic vector of each sentence. 
%The sentence RNN is also a single-layer LSTM with hidden size of 1024 and initial hidden and memory states set to zero.
Specifically, each hidden state $\mathbf{h}^S_t$ is first passed into a linear layer to produce a probability over the two states \{CONTINUE=0, STOP=1\} which determine whether $t$-th sentence is the last sentence. The updated $\mathbf{h}^S_t$ is treated as the topic vector of the sentence.

	\textbf{Word RNN with Language Attention. }
To generate meaningful paragraphs relevant to the image, the model is desired to recognize and describe substantial details such as objects, attributes, and relationships. 
The text phrases of semantic regions that express such local semantics are leveraged by a language attention module to help with the recurrent word generation. For example, the word RNN can conveniently copy precise concepts (e.g., baseball, helmet) from the local phrases. Following the copy mechanism~\cite{gu2016incorporating} firstly proposed in natural language processing, we selectively incorporate the embeddings of local phrases based on the topic vector $\mathbf{h}^S_t$ and preceding word state $\mathbf{h}^w_{t,i-1}, i \in \{1,\dots, \mathbf{N}_t\}$ by the word RNN to generate the next word representation $\mathbf{f}^l_{t,i}$. Since each local phrase $\mathbf{s}^r_j$ semantically relates to respective visual feature $\mathbf{v}_j$, we thus reuse the visual attentive weights $\{a_j\}_{j=1}^M$ to enhance the language attention. Representing each word with an embedding vector $\mathbf{w}^r_{i,j}$, the language representation $\mathbf{f}^l_{t,i}$ for each word prediction at $i$-th step is formulated as
\begin{equation}
\small
\begin{split}
&\mathbf{f}^l_{t,i} = \text{att}^l(\mathbf{S}^r, \mathbf{h}^S_t, \mathbf{h}^w_{t,i-1})\\
=&\sum_{j=1}^{M}\sum_{i'=1}^{\mathbf{N}_j}\frac{\alpha(\mathbf{w}^r_{i',j},\beta(\mathbf{h}^S_t,\mathbf{h}^w_{t,i-1}))}{\sum_{j'=1}^{M}\sum_{i''=1}^{\mathbf{N}_{j'}}\alpha(\mathbf{w}^r_{i'',j'},\beta(\mathbf{h}^S_t, \mathbf{h}^w_{t,i''-1}))} a_j \mathbf{w}^r_{i',j}.
\end{split}
\end{equation}
Given the language representation $\mathbf{f}^l_{t,i}$ as the input at $i$-th step, the word RNN computes a hidden states $\mathbf{h}^w_{t,i}$ which is then used to 
predict a distribution over the words in the vocabulary.
%the word RNN as a single-layer LSTM with hidden size of 512 to generate the words of a sentence. 
%The first input is a special START token, and subsequent inputs are learned embedding vectors for the words of the sentence. The hidden states $\mathbf{h}^w_{t,i}$ at each time step is used to predict a distribution over the words in the vocabulary, and a special END token signals the end of a sentence. 
After obtaining all words of each sentence, these sentences are finally concatenated to form the generated paragraph. 
	
	\subsection{Paragraph Discriminators}
	The paragraph discriminators $\{D^s, D^r\}$ aim to distinguish between real paragraphs and synthesized ones based on the linguistic characteristics of a natural paragraph description. In particular, the sentence discriminator $D^s$ evaluates the plausibility of individual sentences, while the topic-transition discriminator $D^r$ evaluates the topic coherence of all sentences generated so far. With such multi-level assessment, the model is able to generate long yet realistic descriptions.
	Specifically, the sentence discriminator $D^s$ is an LSTM RNN that recurrently takes each word embedding within a sentence as the input, and produces a real-value plausibility score of the synthesized sentence. 
	%is a single-layer LSTM unit with 512 hidden and memory states.
	%It recurrently takes the word embeddings of all words in a sentence, and the output of last recurrent step is be fed into one linear layer to generate one final value. 
	%Following Wasserstein GAN~\cite{arjovsky2017wasserstein}, we drop the sigmoid operation on the final prediction in order to remedy the mode collapse problem in the original GAN~\cite{goodfellow2014generative}. 
	The topic-transition discriminator $D^r$ is another  LSTM RNN which 
	%unit with 512 hidden and memory states, 
	recurrently takes the sentence embeddings of all preceding sentences as inputs and computes the topic smoothness value of the current constructed paragraph description at each recurrent step. 
	%As presented in Section~\ref{sec:objective}, following Wasserstein GAN~\cite{arjovsky2017wasserstein}, our discriminators eliminate the sigmoid function that outputs a probability used in original GAN~\cite{goodfellow2014generative}, and thus do not explicitly try to classify inputs as real or fake.
	
%In this way, the topic coherence after each sentence can be ensured to produce more realistic and reasonable long descriptions. The optimization of both $D^s$ and $D^r$ can be referred in Eqn.(\ref{eq:total}). 

%\quad\\
%Since producing sentences is a discrete sampling process, we perform a ``max-pooling" like procedure to enable the joint optimization of generator and discriminator during back-propagation. That is, the gradients coming from the discriminator are only back-propagated to the select word index, which can be regarded as an action selection step to specify each word.

	\subsection{Implementation Details}\label{sec:imp}
	The discriminators $D^s$ and $D^r$ are both implemented as a single-layer LSTM with hidden dimension of 512. For the generator, the paragraph RNN is a single-layer LSTM with hidden size of 512 and the initial hidden and memory cells set to zero. Similarly, the sentence RNN and word RNN are single-layer LSTMs with hidden dimension of 1024 and 512, respectively. Each input word is encoded as a embedding vector of 512 dimension. The visual feature vector $\mathbf{v}_j$ of each semantic region has dimension of 4096.
	
	The adversarial framework is trained following the Wasserstein GAN (WGAN)~\cite{arjovsky2017wasserstein} in which we alternate between the optimization of $\{D^s, D^r\}$ with Eq.\eqref{eq:d-opt} and the optimization of $G$ with Eq.\eqref{eq:g-opt}. In particular, we perform one gradient descent step on $G$ every time after 5 gradient steps on $\{D^s, D^r\}$. We use minibatch SGD and apply the RMSprop solver~\cite{tieleman2012lecture} with the initial learning rate set to $0.0001$. For stable training, we apply batch normalization~\cite{ioffe2015batch} and set the batch size to $1$ (i.e., ``instance normalization"). In order to make the parameters of $D^s$ and $D^r$ lie in a compact space, we clamp the weights to a fixed box $[-0.01, 0.01]$ after each gradient update. In the semi-supervised setting where only single-sentence captioning for images and standalone paragraph corpus are available, we set the maximal number of sentences in the generated paragraph to $6$ for all images. In the fully-supervised setting, the groundtruth sentence number in each visual paragraph is used to train the sentence-RNN for learning how many sentences are needed. We train the models to converge for 40 epochs. The implementations are based on the public Torch7 platform on a single NVIDIA GeForce GTX 1080.
	%The true sentences for optimizing $D^s$ is sampled from both the paragraph corpus and the annotated captioning for images in semi-supervised setting. During the semi-supervised training, since only first sentences for each image are provided, we set the maximal sentence length $T$ as 6 for all images, and use the groundtruth word length $N_1$ for first sentence generation and predicted $N_t$ when the \emph{END} token is predicted.

	\section{Experiments}
	
	\begin{table*}[!tp]
		\centering\renewcommand\arraystretch{1.1}
		\caption{The performance comparisons with four state-of-the-arts and the variants of our RTT-GAN on paragraph generation in terms of six language metrics. The human performance is included for providing a better understanding of all metrics.}\label{tab:paragraph}
		\begin{tabular}{c|cccccccccccccccccccccp{6em}p{5em}}
			\toprule
			Method & METEOR & CIDEr & BLEU-1 & BLEU-2 & BLEU-3 & BLEU-4\\
			\hline
			Sentence-Concat & 12.05 & 6.82 & 31.11 & 15.10 & 7.56 & 3.98 \\
			\hline
			Template & 14.31 & 12.15 & 37.47 & 21.02 & 12.03 & 7.38\\
			\hline
			Image-Flat~\cite{karpathy2015deep} & 12.82 & 11.06 & 34.04 & 19.95 & 12.20 & 7.71\\
			\hline
			Regions-Hierarchical~\cite{krause2016hierarchical} & 15.95 & 13.52 & 41.90 & 24.11& 14.23 & 8.69\\
			\hline
			{RTT-GAN (Semi- w/o discriminator)} &  {12.35} & {8.96} & {33.82} & {17.40} & {9.01} & {5.88}\\
			{RTT-GAN (Semi- w/o sentence D)} &  {11.22} & {10.04} & {35.29} & {19.13} & {11.55} & {6.02}\\
			{RTT-GAN (Semi- w/o topic-transition D)} &  {12.68} & {12.77} & {37.20} & {20.51} & {12.08} & {6.91}\\
			{RTT-GAN (Semi-)} &  {14.08} & {13.07} & {39.22} & {22.50} & {13.34} & {7.75}\\
			\hline
			{RTT-GAN (Fully- w/o discriminator)} &  {16.57} & {15.07} & {41.86} & {24.33} & {14.56} & {8.99}\\
			{RTT-GAN (Fully-)} &  {17.12} & {16.87} & {41.99} & {24.86} & {14.89} & {9.03}\\
			\hline
			\textbf{RTT-GAN (Semi + Fully)} &  \textbf{18.39} & \textbf{20.36} & \textbf{42.06} & \textbf{25.35} & \textbf{14.92} & \textbf{9.21}\\
			\hline
			Human & 19.22 & 28.55 & 42.88 & 25.68 & 15.55 & 9.66\\
			\bottomrule
		\end{tabular}%
		\vspace{-2mm}
	\end{table*}%
	
		\begin{table}[!tp]
			\centering\renewcommand\arraystretch{1.1}
			\caption{Ablation studies on the effectiveness of key components in the region-based attention mechanism of our RTT-GAN.}\label{tab:attention}
			\begin{tabular}{c|cccccccccccccccccccccp{6em}p{5em}}
				\toprule
				Method & METEOR & CIDEr \\
				\hline
				%Regions-Hierarchical~\cite{krause2016hierarchical} & 15.95 & 13.52 \\
				%\hline
				{RTT-GAN (Fully- w/o phrase att)} &  {16.08} & {15.13} \\
				{RTT-GAN (Fully- w/o att)} &  {15.63} & {14.47} \\
				\hline
				{RTT-GAN (Fully- 10 regions)} &  {14.13} & {13.26} \\
				{RTT-GAN (Fully- 20 regions)} &  {16.92} & {16.15} \\
				\hline
				\textbf{RTT-GAN (Fully-)} &  \textbf{17.12} & \textbf{16.87} \\
				\bottomrule
			\end{tabular}%
			\vspace{-6mm}
		\end{table}%
		
In this section, we perform detailed comparisons with state-of-the-art methods on the visual paragraph generation task in both supervised and semi-supervised settings.
	\subsection{Experimental Settings}
	
	To generate a paragraph for an image, we run the paragraph generator forward until the STOP sentence state is predicted or after $S_{\text{max}} = 6$ sentences, whichever comes first. The word RNN is recurrently forwarded to sample the most likely word at each time step, and stops after choosing the STOP token or after $N_{\text{max}} = 30$ words. We use beam search with beam size 2 for generating paragraph descriptions. Training details are presented in Section~\ref{sec:imp}, and all models are implemented in Torch platform. In terms of the fully-supervised setting, to make a fair comparison with the state-of-the-art methods~\cite{karpathy2015deep,krause2016hierarchical}, the experiments are conducted on the public image paragraph dataset~\cite{krause2016hierarchical}, where 14,575 image-paragraph pairs are used for training, 2,487 for validation and 2,489 for testing. In terms of semi-supervised setting, our RTT-GAN is trained with the single sentence annotations provided in MSCOCO image captioning dataset~\cite{chen2015microsoft} which contains 123,000 images. The image-paragraph validation set is used for validating the semi-supervised paragraph generation. The paragraph generation performance is also evaluated on 2,489 paragraph testing samples. For both fully-supervised and semi-supervised settings, we use the word vocabulary of image-paragraph dataset as~\cite{krause2016hierarchical} does and the 14,575 paragraph descriptions on public image paragraph dataset~\cite{krause2016hierarchical} are adopted as the standalone paragraph corpus for training discriminators. We report six widely used automatic evaluation metrics, BLEU-1, BLEU-2, BLEU-3, BLEU-4, METEOR, and CIDEr. The model checkpoint selection is based on the best combined METEOR and CIDEr score on the validation set. Table~\ref{tab:paragraph} reports the performance of all baselines and our models.

	\subsection{Comparison with the State-of-the-arts}
	
    We obtain the results of all four baselines from~\cite{krause2016hierarchical}. Specifically, \emph{Sentence-Concat} samples and concatenates five sentence captions from the model trained on MS COCO captions, in which the first sentence uses beam search and the rest are samples. \emph{Image-Flat}~\cite{karpathy2015deep} directly decodes an image into a paragraph token by token. \emph{Template} predicts the text via a handful of manually specified templates. And \emph{Region-Hierarchical}~\cite{krause2016hierarchical} uses a hierarchical recurrent neural network to decompose the paragraphs into the corresponding sentences. Same with all baselines, we adopt VGG-16 net~\cite{simonyan2014very} to encode the visual representation of an image. Note that our RTT-GAN and \emph{Region-Hierarchical}~\cite{krause2016hierarchical} use the same dense captioning model~\cite{johnson2016densecap} to extract semantic regions. \emph{Human} shows the results by collecting an additional paragraph for 500 randomly chosen images as~\cite{krause2016hierarchical}. As expected, humans produce superior descriptions over any automatic method and the large gaps on CIDEr and METEOR verify that CIDEr and METEOR metrics align better with human judgment than BLEU scores.
	
		\begin{figure}[!tp]
			\begin{center}
				\includegraphics[scale=0.66]{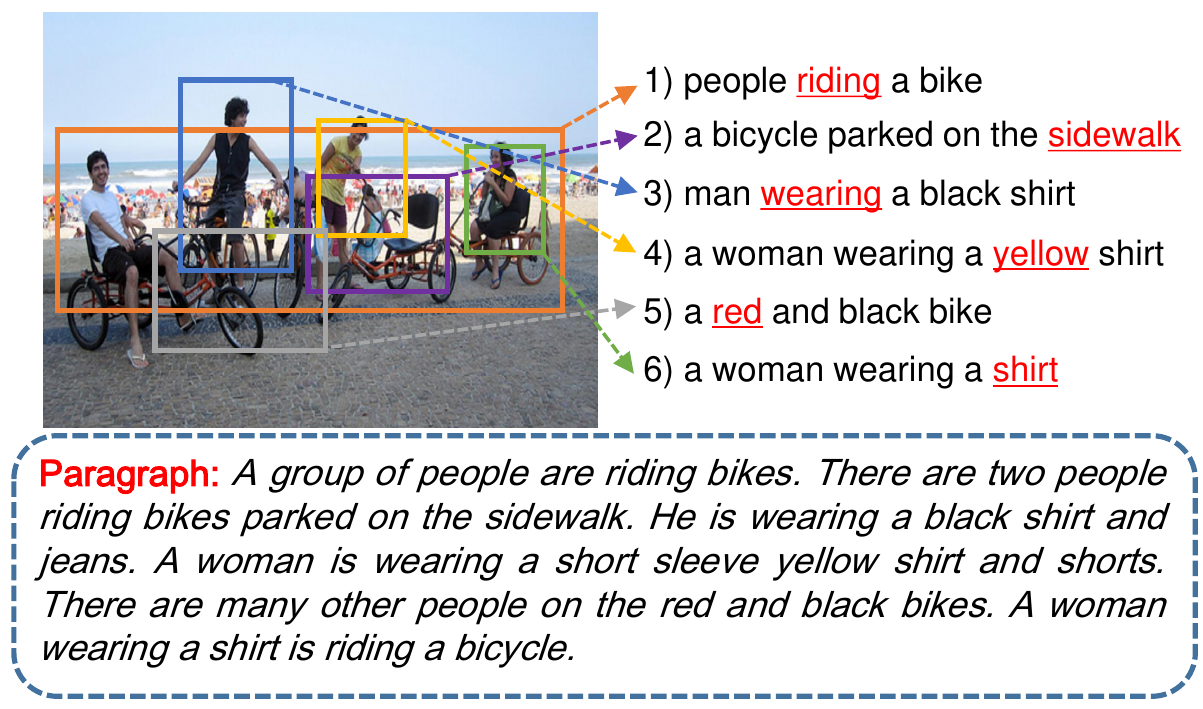}
				\caption{Visualization of our region-based attention mechanism. For each sentence generation, RTT-GAN selectively focuses on semantic regions of interest in the spatial visual attention, and attentively leverage the word embeddings of their local phrases to enhance the word prediction. In the top row, we illustrate the regions with highest attention confidences during the spatial visual attention and its corresponding words (highlighted in red) with highest attention confidences during the language attention in each step.
				} 
				\label{fig:region}
			\end{center}
			\vspace{-8mm}
		\end{figure}

	\textbf{Fully-supervised Setting.} We can see that our \emph{RTT-GAN (Fully-)} model significantly outperforms all baselines on all metrics; particularly, 3.35\% over \emph{Region-Hierarchical} and 5.81\% over \emph{Image-Flat} in terms of CIDEr. It clearly demonstrates the effectiveness of our region-based attention mechanism that selectively incorporate visual and language cues, and the adversarial multi-level discriminators that provide a better holistic, semantic regularization of the generated sentences in a paragraph. The inferiority of \emph{Image-Flat} compared to the hierarchical networks of \emph{RTT-GAN} and \emph{Region-Hierarchical} demonstrates the advantage of performing hierarchical sentence predictions for a long paragraph description. 
	
		\begin{figure*}[!tp]
			\begin{center}
				\includegraphics[scale=0.56]{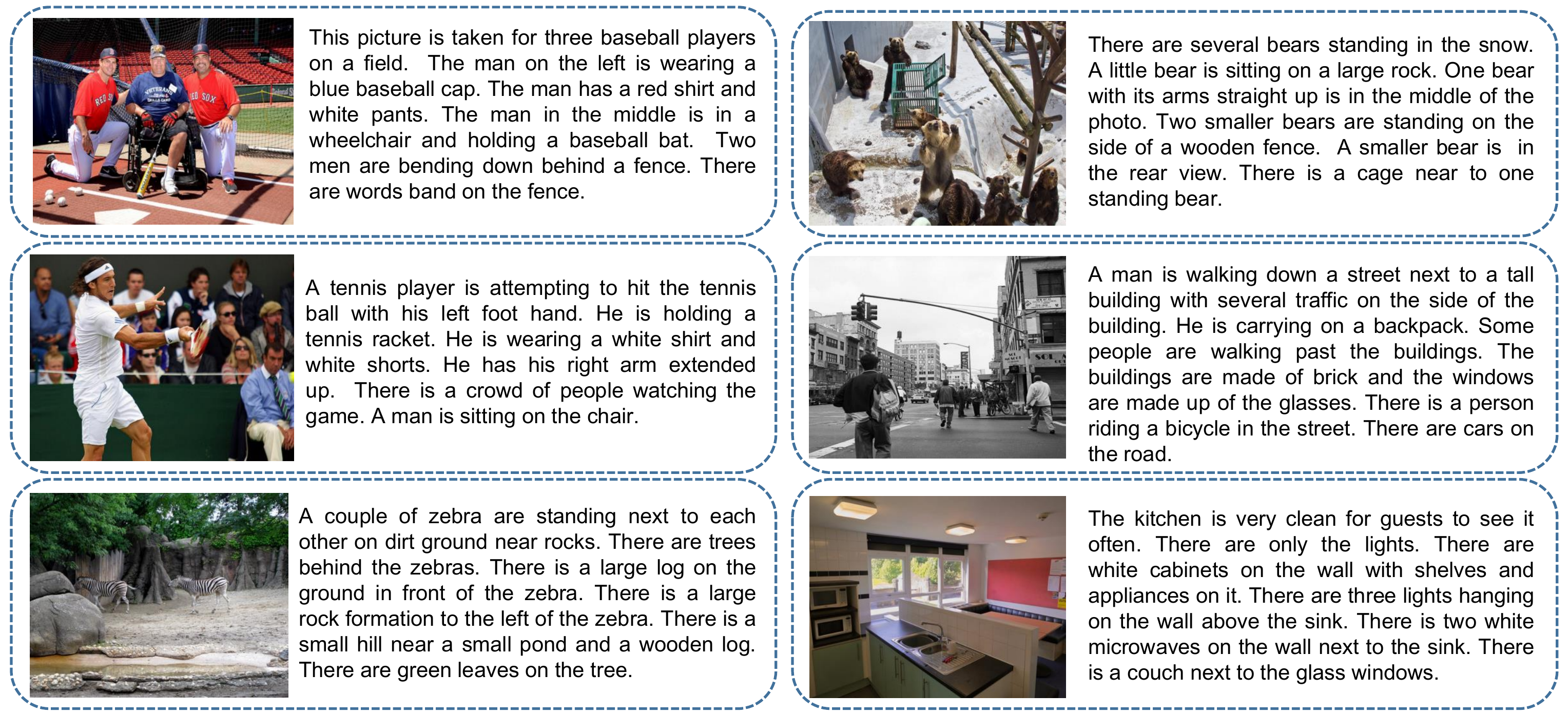}
				\caption{Example paragraph generation results of our model in the semi-supervised setting (\emph{RTT-GAN (Semi-)}). For each image, a paragraph description with all six sentences is generated.
				} 
				\label{fig:semi_results}
			\end{center}
			\vspace{-8mm}
		\end{figure*}
		
	\textbf{Semi-supervised Setting.} The main advantage of our RTT-GAN compared to prior works is the capability of generating realistic paragraph descriptions coordinating with the natural linguistic properties, given only singe sentence annotations. It is demonstrated by the effectiveness of our semi-supervised model \emph{RTT-GAN (Semi-)} that only uses the single sentence annotations of MSCOCO captions, and imposes the linguistic characteristics on the rest sentence predictions using adversarial discriminators that are trained with the standalone paragraph corpus. Specifically, \emph{RTT-GAN (Semi-)} achieves comparable performance with the fully-supervised \emph{Regions-Hierarchical} without using any groundtruth image-paragraph pairs. After augmenting the image paragraph dataset with the synthesized paragraph descriptions by \emph{RTT-GAN (Semi-)}, \emph{RTT-GAN (Semi+Fully)} dramatically outperforms \emph{RTT-GAN (Fully-)} and other baselines, e.g. 6.84\% increase over \emph{Regions-Hierarchical} on CIDEr. We also show some qualitative results of generated paragraphs by our \emph{RTT-GAN (Semi-)} in Figure~\ref{fig:semi_results}. These promising results further verify the capability of our RTT-GAN on reasoning a long description that has coherent topics and plausible sentences without the presence of ground-truth image paragraph pairs.

	\begin{figure*}[!tp]
		\begin{center}
			\includegraphics[scale=0.52]{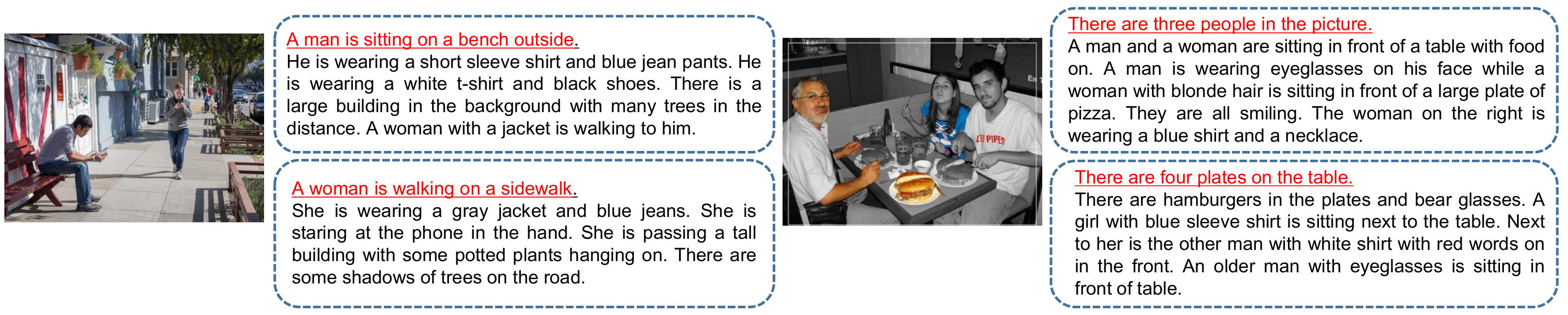}
			\caption{Personalized paragraph generations of our model (i.e. \emph{RTT-GAN (Semi + Fully)}) by manipulating the first sentence. With two different first sentences for each image, our model can effectively generate two distinct paragraphs with different topics. % focuses following the first sentences.
			} 
			\label{fig:personalized}
		\end{center}
		\vspace{-8mm}
	\end{figure*}

	\subsection{The Importance of Adversarial Training} After eliminating the discriminators during the model optimization in both fully- and semi-supervised settings (i.e. \emph{RTT-GAN (Fully- w/o discriminator)} and \emph{RTT-GAN (Semi- w/o discriminator)}), we observe consistent performance drops on all metrics compared to the full models, i.e. 1.80\% and 4.11\% on CIDEr, respectively. \emph{RTT-GAN (Semi- w/o discriminator)} can be regarded as a image captioning model due to the lack of adversarial loss, similar to \emph{Sentence-Concat}. It justifies that the sentence plausibility and topic coherences with preceding sentences are very critical for generating long, convincing stories. Moreover, the pure word prediction loss largely hinders the model's extension to unsupervised or semi-supervised generative modeling. Training adversarial discriminators that explicitly enforce the linguistic characteristics of a good description can effectively impose high-level and semantic constraints on sentence predictions by the generator. %It can be also applied for other vision tasks (e.g. captioning) that attempt to bridge visual and language domain.
	
	Furthermore, we break down our design of discriminators in order to compare the effect of the sentence discriminator and recurrent topic-transition discriminator, as \emph{RTT-GAN (Semi- w/o sentence D)} and \emph{RTT-GAN (Semi- w/o topic-transition D)}, respectively. It can be observed that although both discriminators help bring the significant improvement,  the sentence discriminator seems to play a more critical role by addressing the plausibility of each sentence.

	\subsection{The Importance of Region-based Attention}
	We also evaluate the effectiveness of the spatial visual attention and language attention mechanisms to facilitate the paragraph prediction, as reported in Table~\ref{tab:attention}. \emph{RTT-GAN (Fully- w/o att)} directly pools the visual features of all regions into a compact representation for sequential sentence prediction, like \emph{Region-Hierarchical}. \emph{RTT-GAN (Fully- w/o phrase att)} represents the variant that removes the language attention module. It can be observed that the attention mechanism effectively facilitates the prediction of RTT-GAN by selectively incorporating appropriate visual and language cues. Particularly, the advantages of explicitly leveraging words from local phrases suggest that transferring visual-language knowledges from more fundamental tasks (e.g. detection) is beneficial for generating high-level and holistic descriptions. 
	
	As an exploratory experiment, we investigate generating paragraphs from a smaller number of regions ($10$ and $20$) than $50$ used in previous models, denoted as \emph{RTT-GAN (Fully- 10 regions)} and \emph{RTT-GAN (Fully- 20 regions)}. Although these results are worse than our full model, the performance of using only top 10 regions is still reasonably good. Figure~\ref{fig:region} gives some visualization results of our region-based attention mechanism. For generating the sentence at each step, our model selectively focuses on distinct regions and their distinct corresponding words in local phrases to facilitate the sentence prediction.
		\begin{table}[!tp]
			\centering\renewcommand\arraystretch{1}
			\caption{Human voting results for the plausibility of generated personalized paragraphs by the variants of our RTT-GAN.}\label{tab:personalized}
			\begin{tabular}{c|c|ccccccccccccccccccccp{6em}p{5em}}
				\toprule
				Semi- w/o discriminator & Semi- & Semi + Fully \\
				\hline
				12.6\% & 40.5 \% &  46.9\% \\
				\bottomrule
			\end{tabular}%
			\vspace{-7mm}
		\end{table}%

	\subsection{Personalized Paragraph Generation}
	Different from prior works, our model supports the personalized paragraph generation which produces diverse descriptions by manipulating first sentences. It can be conveniently achieved by initializing the paragraph RNN with the sentence embedding of a predefined first sentence. The generator can sequentially output diverse and topic-coherent sentences to form a personalized paragraph for an image. We present qualitative results of our model in Figure~\ref{fig:personalized}. Some interesting properties of our predictions include
	its usage of coreference in the first sentence and its ability to capture topic relationships with preceding sentences. Given the first sentences, subsequent sentences give
	some details about scene elements mentioned earlier in the description and also connect to other related content. We also report the human evaluation results in Table~\ref{tab:personalized} on randomly chosen 100 testing images, where three model variants are compared, i.e. \emph{RTT-GAN (Semi- w/o discriminator)}, \emph{RTT-GAN (Semi-)}, \emph{RTT-GAN (Semi + Fully)}. For each image, given two first sentences with distinct topics, each model produces two personalized paragraphs accordingly. Regarding to each first sentence of the image, we present three paragraphs by three models in a random order to judges, and ask them to select the most convincing ones. The results in Table~\ref{tab:personalized} indicate that 87.4\% of the judges think the paragraphs generated by the models (i.e. \emph{RTT-GAN (Semi-)}, \emph{RTT-GAN (Semi + Fully)}) that incorporate two adversarial discriminators, look more convincing than those by \emph{RTT-GAN (Semi- w/o discriminator)}.

	\subsection{Extension to Video Domain}
	As in Table~\ref{tab:video}, we also extend our RTT-GAN to the task of video paragraph generation and evaluate it on TACoS-MultiLevel dataset~\cite{rohrbach2014coherent} that contains 185 long videos filmed in an indoor environment, following~\cite{yu2016video}. To model spatial appearance, we also extract 50 semantic regions for the frames in every second. To capture the motion patterns, we enhance the feature representation with motion features. Similar to~\cite{yu2016video}, we use the pre-trained C3D~\cite{tran2014c3d} model on the Sport1M dataset~\cite{karpathy2014large}, which outputs a fixed-length feature vector every 16 frames. We then perform a mean pooling over all features to generate a compact motion representation, which are used as additional inputs in every visual attention step. Our model significantly outperforms all state-of-the-arts, demonstrating its good generalization capability in video domain.

	\begin{table}[!tp]
		\centering\renewcommand\arraystretch{1}
		\caption{Results of video paragraph generation on  TACoS-MultiLevel in terms of BLEU-4, METEOR, CIDEr metrics. }\label{tab:video}
		\begin{tabular}{c|cccccccccccccccccccccp{6em}p{5em}}
			\toprule
			Method & BLEU-4 & METEOR & CIDEr \\
			\hline
			CRF-T~\cite{rohrbach2013translating} & 25.3 & 26.0 & 124.8 \\
			CRF-M~\cite{rohrbach2014coherent} & 27.3 & 27.2 & 134.7 \\
			LRCN~\cite{donahue2015long} & 29.2 & 28.2 & 153.4 \\
			h-RNN~\cite{yu2016video} & 30.5 & 28.7 & 160.2\\
			\hline
			\textbf{RTT-GAN (ours)}  & \textbf{33.8} & \textbf{30.9} & \textbf{165.3}\\
			\bottomrule
		\end{tabular}%
		\vspace{-6mm}
	\end{table}%
	
	\section{Conclusion and Future Work}
	
In this paper, we propose a Recurrent Topic-Transition GAN (RTT-GAN) for visual paragraph generation. Thanks to the adversarial generative modeling, our RTT-GAN is capable of generating diverse paragraphs when only first sentence annotations are given for training. The generator incorporates visual attention and language attention mechanisms to recurrently reason about fine-grained semantic regions. Two discriminators assess the quality of generated paragraphs from two aspects: sentence plausibility and topic-transition coherence. Extensive experiments show the effectiveness of our model in both fully-supervised and semi-supervised settings. In future, we will extend our generative model into other vision tasks that require jointly visual and language modeling.

    {\small
    \balance
		\bibliographystyle{ieee}
		\bibliography{egbib}
	}
	
\end{document}